%% file: main.tex
\documentclass[10pt,twocolumn,letterpaper]{article}

\usepackage{cvpr}              % To produce the CAMERA-READY version
\usepackage{graphicx}
\usepackage{amsmath}
\usepackage{amssymb}
\usepackage{booktabs}
\usepackage{multirow}
\usepackage{pifont}
\usepackage{nicefrac}
\usepackage{comment}
\usepackage{bbding}
\newcommand{\cmark}{\ding{51}}%
\newcommand{\xmark}{\ding{55}}%
\usepackage[pagebackref,breaklinks,colorlinks]{hyperref}

\mathchardef\mhyphen="2D

\usepackage{xcolor}

\usepackage[capitalize]{cleveref}
\crefname{section}{Sec.}{Secs.}
\Crefname{section}{Section}{Sections}
\Crefname{table}{Table}{Tables}
\crefname{table}{Tab.}{Tabs.}

\newcommand{\red}[1]{{\color{red}{#1}}}

\newcommand{\tablestyle}[2]{\setlength{\tabcolsep}{#1}\renewcommand{\arraystretch}{#2}\centering\footnotesize}
\newlength\savewidth\newcommand\shline{\noalign{\global\savewidth\arrayrulewidth
		\global\arrayrulewidth .8pt}\hline\noalign{\global\arrayrulewidth\savewidth}}

\newcommand\blfootnote[1]{%
	\begingroup
	\renewcommand\thefootnote{}\footnote{#1}%
	\addtocounter{footnote}{-1}%
	\endgroup
}

\def\cvprPaperID{840} % *** Enter the CVPR Paper ID here
\def\confName{CVPR}
\def\confYear{2022}

\begin{document}
\title{TransRank: Self-supervised Video Representation Learning \\ via Ranking-based Transformation Recognition}
\author{Haodong Duan$^{1}$ \hspace{5mm} Nanxuan Zhao$^{1,3,4}$\ \Envelope \hspace{5mm} Kai Chen$^{2,5}$ \hspace{5mm} Dahua Lin$^{1,2,3}$\\
$^{1}$The Chinese University of HongKong \hspace{8mm} $^{2}$Shanghai AI Laboratory \\
$^{3}$Centre of Perceptual and Interactive Intelligence \hspace{4mm} $^{4}$University of Bath \hspace{4mm} $^{5}$SenseTime Research }

\maketitle
\input{abstract.tex}
\blfootnote{\Envelope \ Corresponding Author. }
\vspace{-5mm}
\input{intro.tex}

\input{related.tex}
\input{preliminary.tex}

\input{method.tex}

\input{experiment.tex}

\appendix
\input{supp_downstream}

\input{supp_vis}
\input{supp_exp}

%%%%%%%%% REFERENCES
{\small
\bibliographystyle{ieee_fullname}
\bibliography{egbib}
}

\end{document}

%% file: abstract.tex
\begin{abstract}
\vspace{-2mm}
Recognizing transformation types applied to a video clip \emph{(RecogTrans)} is a long-established paradigm for self-supervised video representation learning, which achieves much inferior performance compared to instance discrimination approaches \emph{(InstDisc)} in recent works.
However, based on a thorough comparison of representative RecogTrans and InstDisc methods, 
we observe the great potential of RecogTrans on both semantic-related and temporal-related downstream tasks.
Based on hard-label classification, existing RecogTrans approaches suffer from noisy supervision signals in pre-training.
To mitigate this problem, we developed \textbf{TransRank}, a unified framework for recognizing \textbf{Trans}formations in a \textbf{Rank}ing formulation. 
TransRank provides accurate supervision signals by recognizing transformations relatively, consistently outperforming the classification-based formulation. 
Meanwhile, the unified framework can be instantiated with an arbitrary set of temporal or spatial transformations, demonstrating good generality. 
With a ranking-based formulation and several empirical practices, we achieve competitive performance on video retrieval and action recognition.
Under the same setting, TransRank surpasses the previous state-of-the-art method~\cite{jenni2020video} by \textbf{6.4\%} on UCF101 and \textbf{8.3\%} on HMDB51 for action recognition (Top1 Acc); improves video retrieval on UCF101 by \textbf{20.4\%} (R@1). 
The promising results validate that \emph{RecogTrans} is still a worth exploring paradigm for video self-supervised learning.
Codes will be released at \url{https://github.com/kennymckormick/TransRank}.

\end{abstract}

%% file: intro.tex
\vspace{-4mm}
\section{Introduction}

Effective video representation is of crucial importance for various video understanding tasks, including action recognition~\cite{carreira2017quo, goyal2017something, monfort2019moments}, temporal localization~\cite{zhao2019hacs,caba2015activitynet,zhao2017temporal}, and video retrieval~\cite{xu2016msr,zhou2018towards}. 
To ensure the quality, models pre-trained on large-scale video recognition datasets~\cite{carreira2017quo, monfort2019moments,duan2020omni} have been widely adopted as the initial training point. 
However, labeling such large video datasets is notoriously costly and time-consuming, limiting the growth rate of labeled video datasets and the evolution of supervised video representation.
Considering the infinite supply and the exorbitant annotating cost, learning video representation with self-supervision~\cite{miech2019howto100m,qian2021spatiotemporal,feichtenhofer2021large} has drawn increasing attention.

Relying on pretext tasks, video self-supervised learning (video SSL) can obtain good representation without human annotation. 
The learned representation is then transferred to benefit a series of downstream tasks via finetuning.
Most pretext tasks fall into two broad categories: recognizing transformation types~\cite{jing2018self,benaim2020speednet,jenni2020video} (\textit{RecogTrans}) and instance discrimination~\cite{chen2021rspnet,pan2021videomoco,wang2020self} (\textit{InstDisc}).
RecogTrans pretext tasks aim to classify the transformation applied to video clips. 
The applied transformation can be either spatial (rotation~\cite{jing2018self}, resizing~\cite{li2021video}) or temporal (different playback rates~\cite{benaim2020speednet, jenni2020video, wei2018learning}).
Recently, following the success of contrastive learning in the image domain~\cite{he2020momentum, chen2020simple, grill2020bootstrap}, InstDisc-based pretext tasks gradually become the dominant approach for video self-supervised learning, significantly outperforming RecogTrans-based ones on video downstream tasks, including recognition and retrieval~\cite{soomro2012ucf101,kuehne2011hmdb}.

\begin{figure}[t]
    \vspace{-2mm}
    \centering
    \captionsetup{font=small, position=bottom}
    \captionsetup[subfloat]{font=footnotesize, position=bottom}
    \includegraphics[width=.9\linewidth]{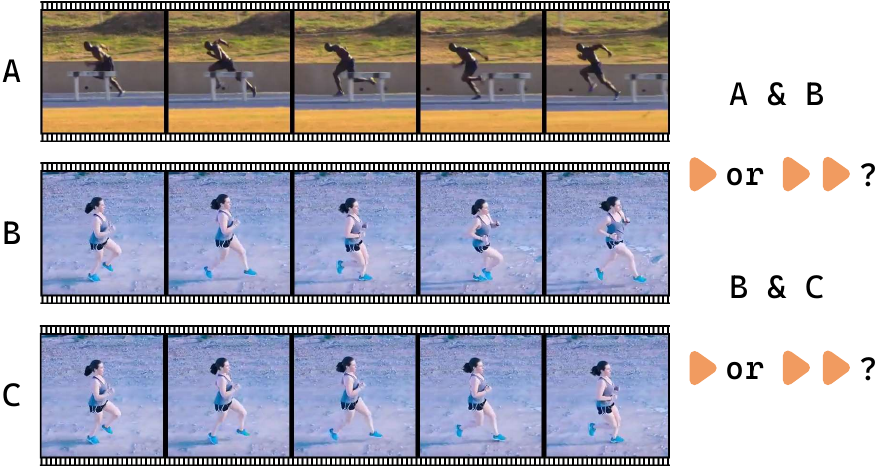}
    \vspace{-2mm}
    \caption{\textbf{Quiz Time! } 
    Two temporal transformations (Normal Speed $1\times$ and Sped Up $2\times$) are applied to 3 clips. 
    In pair (A, B), (B, C), two clips are played with $1\times$, $2\times$ speed, respectively. 
    Can you find which clip is played with $2\times$ speed in (A, B)?
    What about the pair (B, C)? 
    Answers are in the footnote\protect\footnotemark on the next page. }
    \vspace{-5mm}
\label{fig-quiz}
\end{figure}

Rather than directly continuing this trendy research direction, the popularity of InstDisc-based approaches raises several questions that we care about.
First, is there any difference between the representations learned via RecogTrans and InstDisc? 
Is InstDisc-based representation generally more powerful, or do two representations focus on different aspects, having their own merits? 
Besides, what is the primary cause of the inferior performance of RecogTrans? 
Is it because of general limitations of this framework or because of lacking good practices?

To answer these questions, we first conduct a comprehensive study with thorough comparisons on representative methods of RecogTrans and InstDisc.
We find that the representation learned via temporal RecogTrans (\textit{RecogTrans-T}) has some unique properties, distinguishing itself from representations learned via InstDisc and spatial RecogTrans (\textit{RecogTrans-S}). 
The representation learned via RecogTrans-T does not include large amounts of semantic cues, leading to relatively poor performance in downstream settings that directly evaluate the learned representations on semantic-related tasks, like video retrieval and linear evaluation.
Meanwhile, RecogTrans-T shows an impressive capability of temporal modeling and does well across different temporal-related tasks. 
Furthermore, when evaluated on action recognition with a finetuning setting, RecogTrans-T can surpass all alternative SSL methods with a proper finetuning strategy.  
These findings demonstrate the great potential of RecogTrans-T approaches and motivate us to further explore this direction.

\footnotetext{
    A, C are played with $1\times$ speed, B is played with $2\times$ speed. 
    One may give wrong predictions for the pair (A, B) due to their different intrinsic speeds. 
    Meanwhlie, it is easy to find that B is played faster than C. }

RecogTrans suffers from noisy supervision signals caused by the ignorance of video intrinsic properties. 
To this end, we develop \textbf{TransRank}: a unified framework for recognizing \textbf{Trans}formations in a \textbf{Rank}ing formulation.
The core of our TransRank framework is to consider RecogTrans tasks in a relative manner due to the different intrinsic speeds of videos (Figure~\ref{fig-quiz}). 
For example, `boxing' and `running' are much faster than `tai-chi' and `walking' in human perception. 
Even for the same action, the intrinsic speeds can vary a lot when performed by different people (in Figure~\ref{fig-quiz}, the runner in clip-A runs much faster than the runner in clip-B). 
Compared to the classification-based formulation (\textbf{TransCls}), TransRank adopts a more accurate and distinct supervision signal, thus outperforming TransCls consistently across different settings.
Moreover, the ranking formulation is not detrimental to the universality.
TransRank can be instantiated with an arbitrary set of temporal (or spatial) transformations. 
We further conduct an extensive ablation study on choices of the transformation set, good practices during pre-training and finetuning, and evaluate learned representations on diverse downstream tasks. 
Competitive performance on downstream tasks, including video retrieval and action recognition, can be obtained with the ranking-based framework and several good practices.

In summary, we make the following contributions:
 
1) We revisit several SSL approaches based on RecogTrans and InstDisc, demonstrating the \emph{great potential} of RecogTrans in video self-supervised learning.
 
2) We develop a new framework called TransRank, which provides more accurate supervision signals than RecogTrans based on hard-label classification, and can be applied to various temporal and spatial pretext tasks. 

3) With TransRank and several good practices, we improve the RecogTrans-based video SSL to the next level.
Under the same setting, TransRank outperforms a previous state-of-the-art work~\cite{jenni2020video} by \textbf{6.4\%} on UCF101 and \textbf{8.3\%} on HMDB51.  
We achieve decent recognition results (\textbf{90.7\%}, \textbf{64.2\%} on UCF101, HMDB51) with a simple R(2+1)D-18 backbone and visual-only inputs. 
Encouraging results validate that RecogTrans is still worth exploring.

%% file: related.tex
\section{Related Work}

Self-supervised learning is a long-standing problem attracting a lot of research works in different directions, such as image~\cite{noroozi2016unsupervised, gidaris2018unsupervised, doersch2015unsupervised, caron2018deep, oord2018representation}, video~\cite{kim2019self, han2019video, benaim2020speednet, wang2020self, jenni2020video}, and cross-modal~\cite{alayrac2020self, alwassel2019self, patrick2021compositions}. 
We briefly introduce the most related works on self-supervised video representation learning in this section. 
The mainstream methods can be categorized into \textit{RecogTrans} and \textit{InstDisc}.

\textbf{RecogTrans approaches. }
RecogTrans aims to train a model to recognize the applied transformation to data.
Specifically, for video representation learning, the transformations can be either spatial~\cite{kim2019self, jing2018self} or temporal~\cite{benaim2020speednet,jenni2020video, xu2019self, yao2020video, lee2017unsupervised}. 
Spatial RecogTrans extends existing image pretext tasks (\ie, Jigsaw~\cite{noroozi2016unsupervised}, rotation prediction~\cite{gidaris2018unsupervised}) to handle video inputs.
For temporal RecogTrans, one pretext task is the order prediction: frames~\cite{lee2017unsupervised, misra2016shuffle} or clips~\cite{xu2019self} from a video are shuffled, and a model is learned to predict the order. 
Another well-known pretext task is playback style prediction. 
AoT~\cite{wei2018learning} first proposes to predict the arrow of time in videos, which can be regarded as a binary classification of forward or rewind.
SpeedNet~\cite{benaim2020speednet} and PRP~\cite{yao2020video} learn a model for playback rate perception. 
Besides different playing speeds, RTT~\cite{jenni2020video} further introduces more temporal transformations (\eg, periodic, random) to enrich the transformation set for RecogTrans-T. 
Despite the diversified temporal transformations adopted, these works all conduct RecogTrans in a hard-label classification manner, without considering the intrinsic speeds of different videos.
In contrast, our work considers the transformation recognition task in a relative manner, which provides clean and distinct supervision signals in pre-training.

\textbf{InstDisc approaches. }
Learning models to perform instance discrimination is another paradigm in self-supervised learning. 
Following the great success of InstDisc approaches in image representation learning~\cite{he2020momentum, chen2020simple, chen2020big, grill2020bootstrap, chen2021exploring, zbontar2021barlow, caron2020unsupervised,varamesh2020self,brown2020smooth}, 
InstDisc-based video representation learning has attracted wide attention in recent days. 
\cite{wang2020self} first proposes to combine pace prediction and contrastive learning for video representation learning. 
Following works improve vanilla contrastive learning in different ways: VTHCL~\cite{yang2020video} proposes hierarchical contrastive learning to learn visual tempo consistency; VideoMoCo~\cite{pan2021videomoco} adopts adversarial learning as a temporal data augmentation; CoCLR~\cite{han2020self} proposes to learn strong representations via co-training two modalities: RGB and Flow. 
Meanwhile, another line of works~\cite{chen2021rspnet, huang2021ascnet, tao2020selfsupervised, jenni2021time} aims to combine contrastive learning and RecogTrans to learn better representations.
Recent works~\cite{qian2021spatiotemporal, feichtenhofer2021large} show that pure InstDisc approaches can achieve much superior performance on downstream tasks to RecogTrans approaches,  with a strong backbone and a large-scale training set.
Based on thorough comparisons of representative InstDisc and TransRecog methods, we find that representations learned by two method families focus on different aspects: representations learned with InstDisc are \textbf{not} generally more powerful. 
Besides, with good practices, RecogTrans can also achieve competitive downstream performance on semantic tasks like action recognition.

\textbf{Other approaches. }
Besides two main categories, there still exist various approaches that exploit spatiotemporal information in different ways, for example, by future prediction~\cite{han2019video, vondrick2016anticipating}, co-occurrence~\cite{isola2015learning}, or temporal coherence~\cite{lai2020mast, wang2019learning, wang2021unsupervised}. 
Besides, videos are a rich source of multiple modalities.
One can also exploit the rich supervision signals from other modalities, including text~\cite{li2020learning,miech2020end,sun2019learning, alayrac2020self}, audio~\cite{alwassel2019self, korbar2018cooperative, patrick2021compositions}, and optical flow~\cite{gan2018geometry, luo2017unsupervised}. 

%% file: preliminary.tex
\begin{table}[t]
  \vspace{-1mm}
  \captionsetup{font=small, position=top}
  \captionsetup[subfloat]{font=footnotesize, position=top}
  \caption{\textbf{Preliminary results on semantic-related tasks. }}
  \label{tab-preli-semantic}
  \vspace{-2mm}
  \centering
  \subfloat[\textbf{Video retrieval results on UCF101 and HMDB51. \label{tab-preli-retr}}]{
  \resizebox{.95\linewidth}{!}{
  \tablestyle{7pt}{1.15}  
  \begin{tabular}{ccccccc}
      \shline
      \multirow{2}{*}{Method} & \multicolumn{3}{c}{UCF101} & \multicolumn{3}{c}{HMDB51}  \\ 
      & R@1 & R@5 & R@10 & R@1 & R@5 & R@10 \\ \shline
      3D-RotNet~\cite{jing2018self} & 40.8 & 56.8 & 65.0 & 17.5 & 39.1 & 53.1 \\ 
      SpeedNet~\cite{benaim2020speednet} & 24.4 & 39.1 & 48.2 & 12.4 & 30.9 & 43.5 \\ \shline
      SimSiam~\cite{chen2021exploring} & 39.0 & 53.1 & 60.6 & 17.1 & 37.3 & 48.6 \\ 
      MoCo~\cite{he2020momentum} & \textbf{50.1} & \textbf{63.8} & \textbf{71.9} & \textbf{21.8} & \textbf{43.9} & \textbf{57.0} \\ \shline
  \end{tabular}}}
  \vspace{1mm}
  \centering
  \subfloat[\textbf{Video recognition results on UCF101 and HMDB51. \label{tab-preli-recog}}]{
  \resizebox{.95\linewidth}{!}{
  \tablestyle{10pt}{1.15}
  \begin{tabular}{ccccc}
      \shline
      \multirow{2}{*}{Method} & \multicolumn{2}{c}{UCF101} & \multicolumn{2}{c}{HMDB51}  \\ 
      & Linear & Finetune & Linear & Finetune \\ \shline
      3D-RotNet~\cite{jing2018self} & 51.6 & 77.9 & 23.1 & 47.7 \\ 
      SpeedNet~\cite{benaim2020speednet} & 31.6 & \textbf{81.9} & 18.4 & \textbf{51.6} \\ \shline
      SimSiam~\cite{chen2021exploring} & 35.4 & 81.2 & 27.9 & 49.5 \\
      MoCo~\cite{he2020momentum}  & \textbf{63.4} & 79.4 & \textbf{35.3} & 46.2 \\ \shline
  \end{tabular}}}
  \vspace{-5mm}
\end{table}

\section{Preliminary Comparisons of \textit{InstDisc} and \textit{RecogTrans} }

We first conduct pilot experiments to have a preliminary understanding of representations learned by two self-supervised paradigms: \textit{InstDisc} and \textit{RecogTrans}.
We select two representative methods from each paradigm: SpeedNet~\cite{benaim2020speednet} (temporal), 3D-RotNet~\cite{jing2018self} (spatial) for RecogTrans; MoCo~\cite{he2020momentum}, SimSiam~\cite{chen2021exploring} for InstDisc.
We use R3D-18 for all methods and train them on MiniKinetics~\cite{xie2017rethinking} for 200 epochs with strong augmentations in Sec~\ref{sec-aug}\footnote{
  We also perform a basic hyper-parameter search to ensure a suitable setting for each algorithm, \emph{i.e.},
  our MoCo implementation surpasses VideoMoco~\cite{pan2021videomoco} with the same backbone (R3D-18). }. 
In the transferring stage, we evaluate the learned representations on various downstream tasks, which can be categorized as semantic-related tasks and temporal-related tasks. 
We briefly introduce these downstream tasks in this section and leave the detailed setting of each task to the appendix.

\begin{figure}
  \centering
  \vspace{-1mm}
  \captionsetup{font=small, position=bottom}
  \captionsetup[subfloat]{font=footnotesize, position=bottom}
  \includegraphics[width=.9\linewidth]{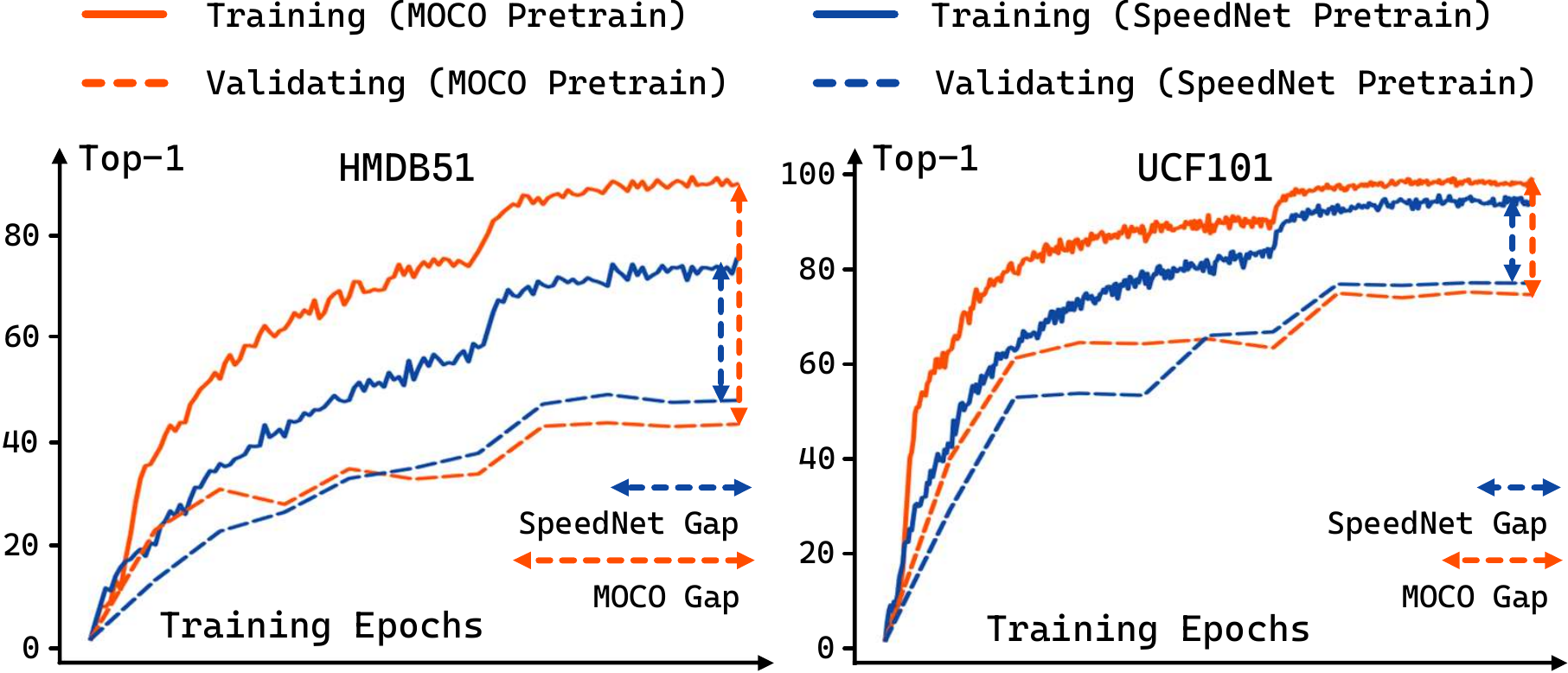}
  \vspace{-2mm}
  \caption{\textbf{The train-validation accuracy gaps on downstream tasks. }
  Initializing with weights pre-trained on RecogTrans-T (i.e., SpeedNet) is less vulnerable to overfitting. }
  \label{fig-trainvalgap}
  \vspace{-4mm}
\end{figure}

\textbf{Semantic-related Tasks. }
Semantic-related tasks are the major downstream tasks used in previous literature, focusing on understanding the semantics of a video clip~\cite{jenni2020video,wang2020self,han2020self}.
There exist three main settings in total:
1). \emph{Nearest-Neighbor Evaluation}, which tests if a query class is present in the top-$k$ retrievals;
2). \emph{Linear Evaluation}, which trains a linear classifier based on the learned representation to classify a novel dataset.
3). \emph{Finetuning}, which finetunes the whole network initialized with the learned weights to train a new classifier. 
In real-world scenarios, finetuning performance is of \emph{greater importance} since better performance on downstream tasks is usually obtained via finetuning.

\textbf{Temporal-related Tasks. } 
Temporal-related tasks focus on recognizing temporal patterns of video clips, which is of crucial importance in video understanding~\cite{zhou2018temporal, lin2019tsm, yang2020temporal}. 
We evaluate the temporal modeling capability of learned representations with three temporal-related tasks:
1). \emph{Motion Type Prediction (Motion)}~\cite{parihar2021spatio}, which categorizes the motion in a video into five pre-defined types (\eg, linear and oscillatory);
2). \emph{Synchronization (Sync)}~\cite{jenni2020video}, which predicts the temporal pattern of two overlapping video clips (\eg, clip-A is ahead of clip-B with $\nicefrac{1}{2}$ overlapping); 
3). \emph{Temporal Order Prediction (Order)}~\cite{jenni2020video}, which determines the temporal order of two non-overlapping clips. 
We demonstrate three tasks in detail in the appendix. 

\begin{table}[t]
  \captionsetup{font=small, position=top}
  \captionsetup[subfloat]{font=footnotesize, position=top}
  \caption{\textbf{Preliminary results on temporal-related tasks. }}
  \label{tab-preli-temporal}
  \vspace{-2mm}
  \centering
  \resizebox{\linewidth}{!}{
  \tablestyle{4pt}{1.15}
  \begin{tabular}{ccccccc}
      \shline
        & Motion & Sync-U & Order-U & Sync-H & Order-H & Avg. Rank \\ \shline
      Random Guess & 20.0 & 14.7 & 50.0 & 14.7 & 50.0  & - \\ \shline
      3D-RotNet~\cite{jing2018self} & 69.1 & 45.8 & 62.6 & 43.3 & 60.3 & 3.4 \\ 
      SpeedNet~\cite{benaim2020speednet} & \textbf{73.3} & 46.5 & \textbf{82.9} & 
      \textbf{46.8} & \textbf{80.6} & \textbf{1.2} \\ 
      \shline
      MoCo~\cite{he2020momentum} & 68.0 & 43.4 & 69.5 & 40.6 & 67.5 & 3.6 \\ 
      SimSiam~\cite{chen2021exploring} & 69.6 & \textbf{48.9} & 75.1 & 45.1 & 71.4 & 1.8 \\ \shline
  \end{tabular}}
  \vspace{-6mm}
\end{table}

We evaluate four SSL methods on all the semantic-related and temporal-related tasks for a comprehensive study. 
We use the official split-1 of UCF101~\cite{soomro2012ucf101} and HMDB51~\cite{kuehne2011hmdb} in all downstream tasks\footnote{Except \emph{Motion}: only HMDB51 is annotated with motion types. }.

Table~\ref{tab-preli-semantic} shows the evaluation results of different representations under three settings for semantic-related tasks.
The performance of video retrieval and linear classification is directly correlated with the quality of the learned representation.
MoCo, thanks to its contrastive pre-training, achieves the best results under the first two settings, while SpeedNet becomes the underdog due to the lack of semantic cues in the learned representation. 
However, once finetuned on downstream datasets, SpeedNet quickly adapts to the action recognition task, outperforming all other methods on UCF101 and HMDB51.
It's also worth noting that the weight pre-trained on SpeedNet is less vulnerable to overfitting when finetuned on downstream recognition tasks (Figure~\ref{fig-trainvalgap}).
For temporal-related tasks (Table~\ref{tab-preli-temporal}), SpeedNet achieves the best results across 4 of 5 evaluation settings, demonstrating its great temporal modeling capability.

So far, we can issue several questions raised before: 
1) The evaluation on diversified downstream tasks demonstrates that representations learned via RecogTrans or InstDisc have their unique benefits: no one is generally more powerful across all evaluation protocols. 
2) There is a good chance that the inferior performance reported in previous RecogTrans works is due to lacking good practices. 
The representation learned by the reimplemented SpeedNet is more powerful compared to the original paper~\cite{benaim2020speednet}, with less training data (MiniKinetics \emph{vs.} K400) and an inferior backbone (R3D-18 \emph{vs.} S3D-G). 
The improvement is due to the better practices we adopted, like strong augmentations or large finetuning learning rates.

Preliminary experiments have shown good representation learning capability and great potential of RecogTrans-T.
However, SpeedNet, as a possible instantiation, still has plenty of room for improvement.
First, in SpeedNet, clips are directly categorized as Normal Speed or Sped Up without considering the intrinsic speeds of different videos.
Besides, with only two temporal transformations, it is highly likely that SpeedNet did not exploit the full capacity of RecogTrans-T. 
In this work, we develop a unified framework \textbf{TransRank} to further realize the potential. 
TransRank can achieve outstanding transfer learning performance on downstream tasks, providing substantial evidence for the potential of RecogTrans.

%% file: method.tex
\begin{figure*}[t]
    \centering
    \vspace{-2mm}
    \captionsetup{font=small, position=bottom}
    \captionsetup[subfloat]{font=footnotesize, position=bottom}
    \includegraphics[width=.9\linewidth]{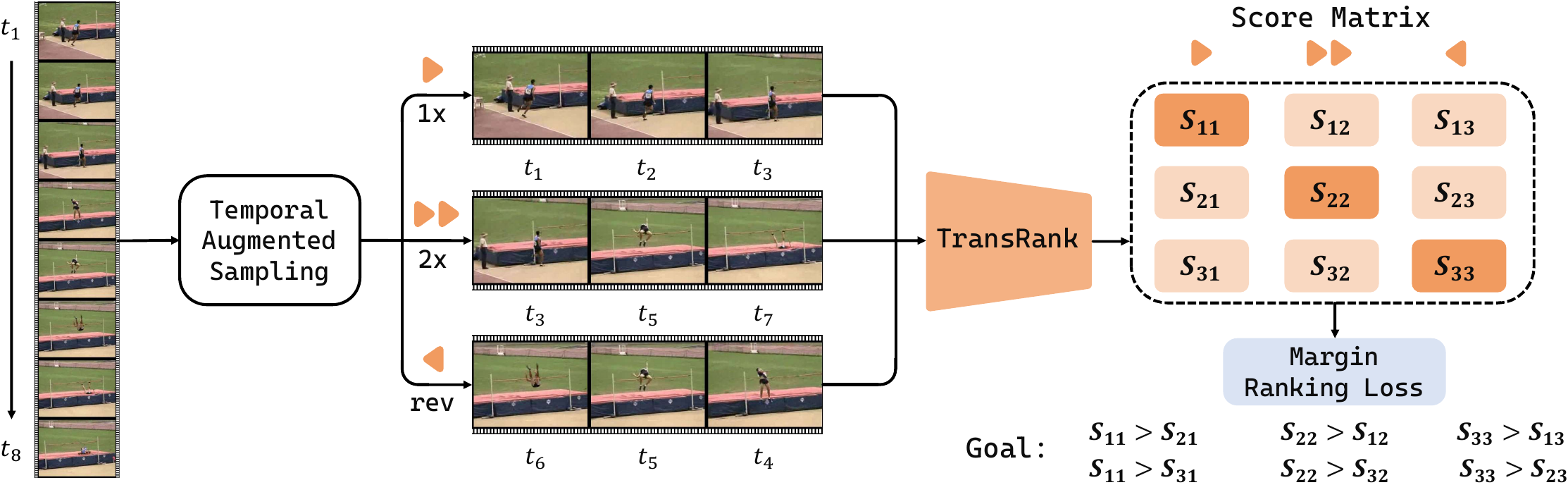}
    \vspace{-3mm}
    \caption{\textbf{The TransRank Framework. } 
        In this example, three clips are sampled from a video, with three different temporal transformations: $1\times$, $2\times$, and $\mathrm{rev}$ applied, respectively. 
        For each clip, TransRank predicts three scores, corresponding to three transformations. 
        A margin ranking loss is applied to the $3\times 3$ score matrix.}
    \vspace{-4mm}
\label{fig-TransRank}
\end{figure*}

\section{TransRank: Ranking-based RecogTrans}

TransRank well improves two drawbacks mentioned above.
With a ranking-based formulation, TransRank considers the intrinsic properties (like speeds) of videos and 
the per-video basis difficulty of distinguishing different transformations 
(\emph{e.g.}, it can be easy, hard, or impossible to distinguish forward and rewind for different videos), 
thus providing cleaner supervision signals. 
Meanwhile, it can be instantiated with an arbitrary set of transformations, thus retaining the generality.

\subsection{Learning Framework } 
TransRank recognizes transformations relatively (\emph{e.g.}, a video played with $2\times$ speed is faster than the same video played with $1\times$ speed), 
and supports various temporal (or spatial) transformations. 
A deterministic version of TransRank is illustrated in Figure~\ref{fig-TransRank}, which samples three clips from each video and applies three different transformations to each of them. 
More generally, TransRank can be instantiated with $M$ different temporal transformations $\{T_{1}, ..., T_{M}\}$.
In the training stage, it samples $N$ different clips $\{c_1, ..., c_N\}$ from a video, and apply a temporal transformation $T_{t_i}, t_i \in \{1, ..., M\}$ deterministically or stochastically to each $c_i$.
For each clip $c_i$, TransRank predicts a score vector $S_{i}=[s_{i1}, ..., s_{iM}]$, $s_{ij}$ denotes the confidence score of the clip $c_i$ being transformed by $T_j$. 
On top of the score matrix $S \in \mathbb{R}^{N \times M}$, TransRank adopts a margin ranking loss to learn the transformation recognition task: 
\vspace{-1mm}
\begin{equation}
\vspace{-2mm}
\small
\mathcal{L_{T\mhyphen MR}} = \frac{1}{\sum\limits_{i, j} \mathbf{1}_{t_i\neq t_j}(i, j) } \sum\limits_{i,j,t_i\neq t_j} \max (0, s_{jt_i}-s_{it_i}+m), 
\end{equation}
where $m$ indicates the margin, which is a hyper-parameter and $\mathbf{1}_{t_i\neq t_j}$ is an indicator that equals to 1 if $t_i\neq t_j$, otherwise 0. 
The intuition is, for two clips $c_i, c_j$ sampled from a same video with $T_{t_i}, T_{t_j}$ applied, respectively, the score $s_{it_i}$ should be larger than $s_{jt_i}$, and vice versa. 
Meanwhile, for a classification formulation, the cross-entropy loss is used: 
\vspace{-2mm}
\begin{equation}
\vspace{-2mm}
\small
\mathcal{L_{T\mhyphen CE}} = \frac{1}{N} \sum_{i} - log (exp(s_{it_i}) / \sum\limits_{j=1}\limits^M exp(s_{ij})). 
\end{equation}

\begin{figure}[t]
    \centering
    \vspace{-4mm}
    \captionsetup{font=small, position=bottom}
    \captionsetup[subfloat]{font=footnotesize, position=bottom}
    \includegraphics[width=.9\linewidth]{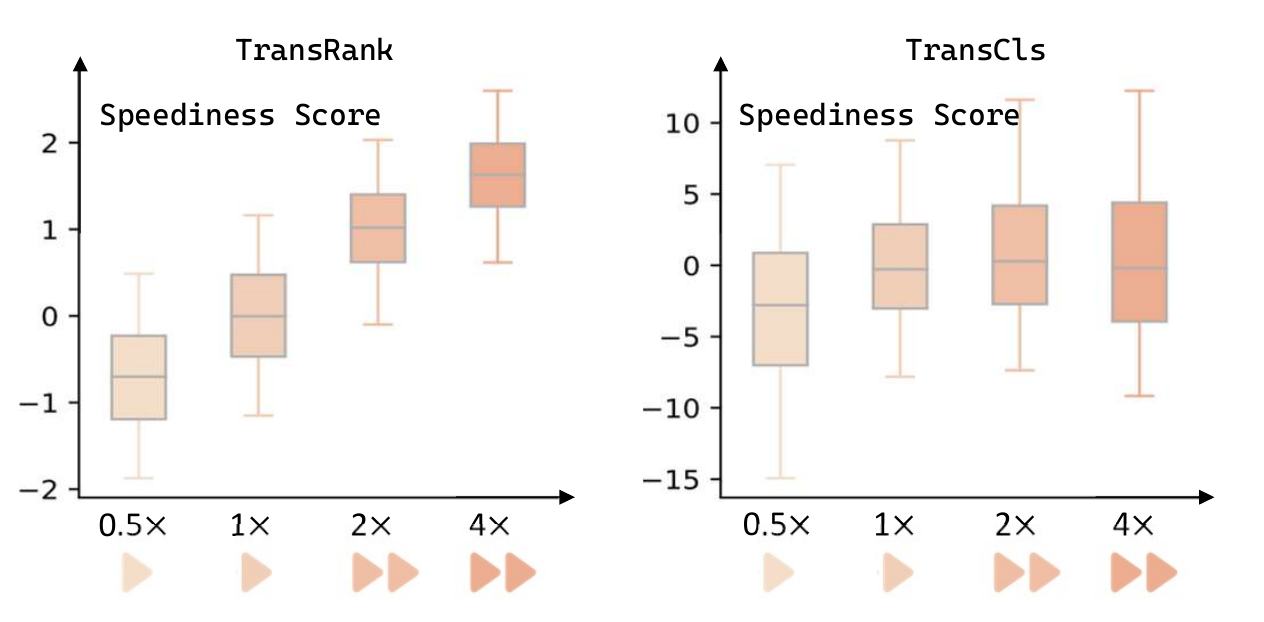}
    \vspace{-3mm}
    \caption{\textbf{TransRank vs. TransCls: The Speediness Distribution on MiniKinetics validation dataset.} 
    A stronger transformation (e.g., $4\times$) should have a larger speediness score. 
    TransRank can successfully reflect this, while TransCls failed to achieve it. \protect\footnotemark}
    \vspace{-4mm}
\label{fig-speediness}
\end{figure}

Intuitively, representations learned via TransRank are of better temporal modeling capability: the supervision signal is less noisy with the intrinsic speeds of videos considered.
For a better illustration, we conduct a toy experiment and show the results in Figure~\ref{fig-speediness}.
We train TransRank and TransCls separately with two temporal transformations: Normal Speed ($T_1: 1\times$) and Sped Up ($T_2: 2\times$) on MiniKinetics\footnote{
For MiniKinetics videos, 4 transformations $0.5\times$, $1\times$, $2\times$, $4\times$ sample 1 frame per 1, 2, 4, 8 frames to form a clip, respectively. 
}. 
Then for each testing clip $c_i$, two scores $s_{i1},s_{i2}$ will be predicted per model. 
To verify whether the model is aware of the temporal variance, during testing, we also send clips with two unseen transformations $0.5\times$ and $4\times$ into models for predictions.
The underlying idea is that the predicted scores should reflect the scales of transformations adequately. 
Specifically, we define $s_{i2}-s_{i1}$ as the \emph{speediness score}, and it should be larger with a more drastic transformation.
With ranking-based training, TransRank can generalize to unseen transformations $0.5\times$ and $4\times$, predicting the speediness score accurately. 
Instead, the new temporal transformations cannot be handled by TransCls.

\subsection{On Building the Temporal Transformation Set }
\label{sec-ttrans}

TransRank can be instantiated with an arbitrary set of temporal transformations. 
However, not every transformation will be helpful in representation learning. 
We briefly introduce the temporal transformations we used and discuss the characteristic that a `good' transformation needs to own.

A major family of transformations used in previous works~\cite{benaim2020speednet,chen2021rspnet,jenni2020video,wang2020self} is speed-related ones. 
Different speed-related transformations artificially change the frame rate when sampling clips ($l$ consecutive frames form a clip) from a video. 
We consider three speed-related transformations $\{1\times, 2\times, 4\times\}$ with a base frame interval 2: for transformation $n \times$, the index sequence of frames in an $l$-frame clip is $\rho + [0, 2n, 2\cdot 2n, ..., (l-1)\cdot 2n]$, where $\rho$ is a random offset.

Inspired by the AoT~\cite{wei2018learning} task, we also use another temporal transformation $\mathrm{rev}$ in TransRank, which performs frame sampling reversely. 
$\mathrm{rev}$ can also be instantiated with different speeds, namely $\mathrm{rev}$-$n\times$: the index sequence of frames in a $\mathrm{rev}$-$n\times$ clip is $\rho + [(l-1)\cdot 2n, (l-2)\cdot 2n, ..., 2n, 0]$. 
We abbreviate $\mathrm{rev}$-$1\times$ as $\mathrm{rev}$.

It's worth noting that \textbf{not all} transformations are useful in TransRank. 
One principle is, the model needs to analyze as many frames as possible to recognize a transformation. 
If a transformation can be recognized by watching a small proportion of frames, the model may take the shortcut and fail to learn a good representation for the whole clip. 
We provide two examples of such `bad' transformations, namely $\mathrm{shuffle}$ (frame indices are randomly permuted)  and $\mathrm{palindrome}$ ($\rho + [0, 2\cdot 2n, ..., (l-2)\cdot 2n, (l-1)\cdot 2n, ..., 2n]$), for which critical cues for recognizing the transformation can be obtained by watching a small part of the clip.
In experiments, we find recognizing such transformations is much easier, while adding them to the transformation set leads to performance degradation on downstream tasks.

\footnotetext{
Speediness scores are normalized so that the average score of $1\times$ clips is 0, and the average of $2\times$ clips is 1. 
Whiskers denote 5\%, 95\% quantiles; boxes denote 25\%, 75\% quantiles; middle lines in boxes denote the median. 
Please find class-specific visualization in the appendix.
}

\subsection{The Generality of TransRank }
TransRank is a general framework, not restricted to temporal transformations. 
We take two spatial transformations as examples to show: 
1). How to apply TransRank to recognize spatial transformations? 
2). Which kind of spatial transformation can benefit from the ranking formulation?

\textbf{Estimate Aspect Ratio. }
RandomResizedCrop is often adopted for data augmentation, which crops a region of a random size and aspect ratio from the original clip and resizes it to a target size. 
Inspired by V3S~\cite{li2021video}, we propose to exploit the original aspect ratio of a cropped clip as a supervision signal for video SSL. 
However, videos may have different intrinsic aspect ratios, just like they have different intrinsic speeds. 
Formulating the aspect ratio estimation as a classification or regression problem is unreasonable, while TransRank can be a suitable framework. 
For $N$ clips $\{c_1, ..., c_N\}$ sampled from a video with original aspect ratios $\{r_1, ..., r_N\}$, we estimate a ratio score $r'$ for each clip (lower $r' \rightarrow$ smaller $r$). 
A margin ranking loss is used to ensure the ranking of $r'$ to be the same as $r$: 
\vspace{-1mm}
\begin{equation}
\small
\mathcal{L_{S\mhyphen MR}} = \frac{2}{N\cdot (N-1)} \sum_{i, j, r_i < r_j} \left[\max (0, r'_i - r'_j + m) \right]. 
\vspace{-1mm}
\end{equation}

\textbf{Estimate Rotation. }
Estimating rotations for images or videos is a well-known pretext task in SSL~\cite{gidaris2018unsupervised,jing2018self}. 
Each image or video clip is randomly rotated with four different degrees (0$^{\circ}$, 90$^{\circ}$, 180$^{\circ}$, 270$^{\circ}$), and a network is trained to recognize the applied rotation. 
This task can also be jointly trained with temporal TransRank.  
However, the ranking formulation does not lead to additional benefits over TransCls, since the rotation category is a more distinct supervision signal compared to the playback rate.

\textbf{Final Loss.} 
By jointly optimizing the temporal TransRank objective $\mathcal{L_{T\mhyphen MR}}$ and the spatial TransRank objectives $\mathcal{L_{S\mhyphen MR_{\mathrm{i}}}}$, the final loss is:
\vspace{-1mm}
\begin{equation}
\small
\mathcal{L} = \mathcal{L_{T\mhyphen MR}} + \sum_i \lambda_i \mathcal{L_{S\mhyphen MR_{\mathrm{i}}}}, 
\vspace{-1mm}
\end{equation}
$\lambda_i$ is the loss weight for the $i_{th}$ spatial pretext task.

% \subsection{Implementation Details }

\subsection{Data Augmentations }
\label{sec-aug}

To consider an extreme case: if two clips ($a: 1\times, b: 2\times$) are sampled from a video with no data augmentation, they can be easily distinguished with no knowledge required.
One may just find the $3_{rd}$ frame of $a$ and the $2_{nd}$ frame of $b$ are the same (and the $1_{st}$ frame of $a$ and $b$ are the same). 
When sampling multiple clips from a video, we perform strong spatiotemporal augmentations to prevent TransRank from taking shortcuts and secure a useful representation. 

\textbf{Temporal Augmentations. } 
Two temporal augmentations are applied when sampling clips from a video: 
1) When selecting indices for a clip, to introduce variability in the time domain, we add random jitter to the frame interval. 
We multiply it by a random factor sampled from a uniform distribution $\left[0.8, 1.2\right]$ (\eg, $1\times$: $0.8 - 1.2\times$, $2\times$: $1.6 - 2.4\times$, still distinguishable). 
2) Multiple clips are uniformly sampled from the entire video, with different clip offsets.

\textbf{Spatial Augmentations. }
We apply \emph{clip-wise} strong spatial augmentations to multiple clips sampled from the same video.
Take RandomResizedCrop as an example: clips in a video are cropped with different random regions. 
Other spatial augmentations we used include RandomGrayScale and RandomColorJitter. 
For joint training with rotation estimation, we further apply RandomRotate, which randomly rotates a video clip with four degrees.

%% file: experiment.tex
\section{Experiments}

\subsection{Implementation Details. }

\textbf{Network Architectures and Inputs.}
We mainly consider two backbones: R3D-18~\cite{hara2018can} and R(2+1)D-18~\cite{tran2018closer}, used frequently in previous video SSL literature.
$N$ clips from a video are fed into the backbones, followed by global average pooling, to extract $N$ 512-d feature vectors. 
Based on extracted features, a temporal head is adopted to predict $M$-d confidence scores for $M$ temporal transformations. 
For optional spatial pretext tasks, one spatial head is used to predict scores for each task. 
The temporal and spatial heads can be instantiated with fully connected layers or multi-layer perceptrons. 
Our implementation is based on the action recognition codebase MMAction2~\cite{2020mmaction2}. 
More details on pre-training and finetuning are in the appendix.

\textbf{Datasets. } We use four datasets in experiments: 
1). \emph{MiniKinetics}\cite{xie2017rethinking} is a subset of the widely used action recognition dataset Kinetics400. 
It consists of 200 action classes and 76K training videos; each lasts around 10 seconds.
We use the training split of MiniKinetics as the pre-training dataset and \emph{K200} to denote MiniKinetics in experiments.
2). \emph{SthV1}\cite{goyal2017something} is a large collection of labeled video clips showing humans performing pre-defined basic actions. 
It consists of 86K training videos and 11K validation videos covering 174 action categories. Each clip lasts around 2 to 6 seconds. 
We treat SthV1 as another pre-training dataset and also adopt it as a downstream recognition task in some experiments using K200 for pre-training. 
3). \emph{UCF101}\cite{soomro2012ucf101} \& \emph{HMDB51}\cite{kuehne2011hmdb} are two relatively small action recognition datasets, consisting of 13K videos of 101 classes and 7K videos of 51 classes, respectively. 
Both datasets are divided into three official train/test splits. 
We use the split-1 for ablation study and report the average performance over three splits for comparison with the state-of-the-art.

\subsection{Ablation Study}
\label{sec-ablation}

By default, we initialize TransRank-T with temporal-transformation (T-Trans) set $\{1\times, 2\times, \mathrm{rev}\}$ and R3D-18, adopt K200 as the pre-training dataset.  
All heads are instantiated with fully-connected layers unless specified. 

\begin{table}[t]
    \captionsetup{font=small, position=top}
    \captionsetup[subfloat]{font=footnotesize, position=top}
    \caption{\textbf{The effect of different temporal transformations on downstream action recognition tasks. }}
    \label{tab-ttrans-set}
    \vspace{-2mm}
    \resizebox{\linewidth}{!}{
		\tablestyle{6pt}{1.2}
    \begin{tabular}{lcccc}
    \shline
     & \multicolumn{2}{c}{\textbf{TransRank (Ours)}} & \multicolumn{2}{c}{TransCls} \\
    T-Trans Set & UCF101 & HMDB51 & UCF101 & HMDB51 \\ 
    \shline
    \{$1\times, 2\times$\} & 81.3 & 51.4 & 80.5 & 50.8 \\ 
    \shline
    \{$1\times, 2\times, \mathrm{rev}$\} & 84.1 & 54.5 & 83.3 & 52.7 \\ 
    \shline
    \{$1\times, 2\times, \mathrm{rev}, \mathrm{palindrome}$\} & 82.9 & 54.7 & 82.4 & 54.0 \\ 
    \{$1\times, 2\times, \mathrm{rev}, \mathrm{shuffle}$\} & 83.8 & 53.6 & 83.0 & 53.6 \\ 
    \{$1\times, 2\times, \mathrm{rev}, 4\times$\} & 85.1 & 56.3 & 84.2 & 54.3 \\ 
    \{$1\times, 2\times, \mathrm{rev}, \mathrm{rev}\mhyphen 2\times$\} & 85.3 & 54.9 & 84.9 & 53.6 \\ 
    \shline
    \end{tabular}}
    \vspace{-5mm}
\end{table}

\textbf{The effect of different T-Trans sets.}
To evaluate the advantage of the ranking formulation, we first compare the downstream performance of TransRank and TransCls instantiated with the same T-Trans set. 
For both frameworks, we start with the T-Trans set $\{1\times, 2\times\}$ (same as SpeedNet), and then add another major transformation $\mathrm{rev}$ and two auxiliary transformations $4\times, \mathrm{rev}\mhyphen 2\times$. 
We show results in Table~\ref{tab-ttrans-set}.
The downstream recognition accuracies increase monotonically under both frameworks as the T-Trans set gets larger, validating the importance of having diverse transformations for RecogTrans-T to learn good representations.
Besides, TransRank outperforms TransCls consistently across different temporal transformation sets, demonstrating the effectiveness of the TransRank framework.

We further add two transformations: $\mathrm{palindrome}$ and $\mathrm{shuffle}$ to the T-Trans set $\{1\times, 2\times, \mathrm{rev}\}$, to validate the design principle of transformation (Sec~\ref{sec-ttrans}).
We observe performance drops in Table~\ref{tab-ttrans-set} and also find that TransRank can easily recognize these two transformations (with Top-1 accuracy $> 99.5\%$) during pre-training. 
The findings are consistent with our assumption in Sec~\ref{sec-ttrans} that the model may take the shortcut and fail to learn a good representation as it can achieve the task by watching a small portion of frames.

\begin{table}[t]
  \vspace{-1mm}
  \captionsetup{font=small, position=top}
  \captionsetup[subfloat]{font=footnotesize, position=top}
  \caption{\textbf{The effect of spatial and temporal augmentations.}}
  \label{tab-ablaug}
  \vspace{-2mm}
  \centering 
  \resizebox{.8\linewidth}{!}{
  \tablestyle{7pt}{1.15}
  \begin{tabular}{cccc}
  \shline
  \multicolumn{2}{c}{Experimental Setup} & \multicolumn{2}{c}{Downstream Tasks} \\
  \shline
  Spatial Augs. & Temporal Augs. & HMDB51 & UCF101 \\ 
  \shline
  \xmark & \xmark & 50.4 & 80.2 \\
  \xmark & \cmark & 50.8 & 80.8 \\
  \cmark & \xmark & 54.2 & 83.8 \\
  \cmark & \cmark & \textbf{54.5} & \textbf{84.1} \\
  \shline
  \end{tabular}}
\end{table}

\begin{figure}
  \vspace{-1mm}
  \captionsetup{font=small, position=bottom}
  \captionsetup[subfloat]{font=footnotesize, position=bottom}
  \includegraphics[width=\linewidth]{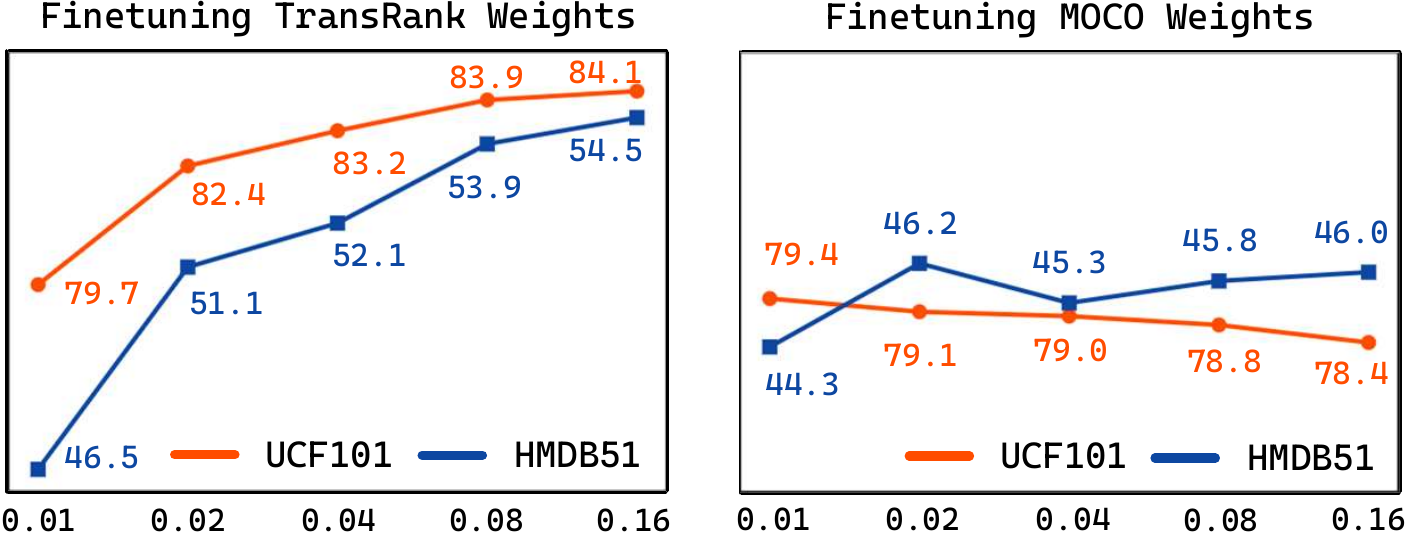}
  \vspace{-6mm}
  \caption{\textbf{The effect of LRs on finetuning SSL models.} 
  The finetuning LR ranges from 0.01 to 0.16 with a batch size of 128. 
  Finetuning with a large LR is essential for TransRank to work well (x-axis: LR, y-axis: Top1 Acc). 
  }
  \label{fig-lrmatters}
  \vspace{-6mm}
\end{figure}

\textbf{Good practices for pre-training \& finetuning. }
We adopt several good practices for pre-training and finetuning of TransRank, including strong augmentations (spatial \& temporal) and large finetuning learning rates. 
We first ablate the effect of strong augmentations. 
Table~\ref{tab-ablaug} demonstrates that both spatial and temporal augmentations contribute to learning good representations for transferring, while the improvement from strong spatial augmentations is more significant. 
Besides, finetuning with a large learning rate (LR) is critical for TransRank to succeed on downstream recognition tasks. 
We finetune TransRank and MoCo with a wide range of the initial LR. 
Figure~\ref{fig-lrmatters} shows that for TransRank, the downstream recognition performance may drastically deteriorate with a small initial LR.
The Top-1 accuracy drops 4.4\% and 8\% on UCF101 and HMDB51, respectively, when using 0.01 as the initial LR. 
For the InstDisc approach MoCo, the downstream performance is much less vulnerable to changes of LR.

\begin{table}[t]
  \vspace{-1mm}
  \captionsetup{font=small, position=top}
  \captionsetup[subfloat]{font=footnotesize, position=top}
  \caption{\textbf{Extending TransRank on spatial transformations.} 
  We jointly train TransRank-T with two spatial tasks. 
  For both tasks, our ranking formulation achieves better or comparable performance than vanilla classification formulation.}
  \label{tab-transrankST}
  \vspace{-2mm}
  \centering 
  \resizebox{.7\linewidth}{!}{
  \tablestyle{10pt}{1.2}
  \begin{tabular}{ccc}
  \shline
  Method & UCF101 & HMDB51 \\
  \shline 
  TransRank-T & 84.1 &  54.5  \\
  \shline
  + Aspect (Cls) & 84.1 & 53.5  \\
  + Aspect (Ours) & \textbf{84.8} & \textbf{56.6}  \\
  \shline
  + Rotation (Cls) & 85.2 & \textbf{57.1}  \\
  + Rotation (Ours) & \textbf{85.3} & 56.8 \\
  \shline
  \end{tabular}}
\end{table}

\textbf{TransRank + spatial transformations. } 
TransRank is a general framework and can be further extended to spatial transformations. 
We formulate aspect ratio estimation and rotation estimation to fit the TransRank framework and jointly train the spatial tasks with TransRank-T.
The loss weight $\lambda$ for each spatial task is 0.5. 
We compare TransRank and TransCls for each spatial task in Table~\ref{tab-transrankST}, evaluating their finetuning performance on action recognition. 
For aspect ratio estimation, TransRank outperforms TransCls by a large margin. 
Since each video may have its own aspect ratio, it is unreasonable to classify with a hard label.
Thus, there is no performance gain after adding this task on TransCls.
Since the supervision signal is distinct and less ambiguous for rotation estimation, both formulations lead to good finetuning performance.

\begin{table}[t]
  \vspace{-1mm}
  \captionsetup{font=small, position=top}
  \captionsetup[subfloat]{font=footnotesize, position=top}
  \caption{\textbf{Improve the feature quality with MLP heads. } 
        The feature quality can be significantly improved by replacing the fc head with a 2-layer MLP (TransRank-ST includes two spatial tasks). }
  \label{tab-featquality}
  \vspace{-2mm}
  \centering 
  \resizebox{.95\linewidth}{!}{
  \tablestyle{6pt}{1.2}
  \begin{tabular}{ccccccc}
  \shline
  Method & w. MLP & Order & Sync & R@1 & R@5 & R@10 \\
  \shline 
  TransRank-T & \xmark & 89.0 & 52.1 & 22.0 & 44.1 & 54.4 \\
  TransRank-T & \cmark & \textbf{92.0} & \textbf{57.4} & \textbf{32.5} & \textbf{54.0} & \textbf{65.0} \\
  \shline
  TransRank-ST & \xmark & 82.3 & 53.7 & 39.2 & 59.1 & 70.0 \\
  TransRank-ST & \cmark & \textbf{83.3} & \textbf{58.4} & \textbf{51.1} & \textbf{70.4} & \textbf{78.3} \\
  \shline
  \end{tabular}}
  \vspace{-5mm}
\end{table}

\textbf{Improve feature quality with MLP heads. }
With finetuning, TransRank achieves competitive results on downstream tasks.
However, the experiments show that RecogTrans approaches are of inferior performance under settings directly evaluating the learned features, \emph{i.e.}, video retrieval. 
We ascribe the deficiency to the head design in previous works~\cite{jenni2020video, benaim2020speednet}: 
using a single fully-connected (FC) layer to predict the confidence for transformations. 
We replace the FC head with a 2-layer MLP (128 hidden channels). 
Table~\ref{tab-featquality} shows that the simple modification drastically improves the feature quality. 
On UCF101, the Top-1 recall increases by 10\% for both TransRank-T and TransRank-ST.
Besides, the improvements of two temporal-related tasks \textit{Sync} and \textit{Order} are also significant. 
We also evaluate the pre-trained models under the finetuning setting, 
but have not observed any significant difference (FC head \emph{vs.} MLP head).

\begin{table}[t]
    \vspace{-1mm}
    \captionsetup{font=small, position=top}
    \captionsetup[subfloat]{font=footnotesize, position=top}
    \caption{\textbf{Comparisons with other video self-supervised learning methods on UCF101 and HMDB51. }
        Methods with * include InstDisc in pre-training.  
        Methods with \# use heavy backbones.  
        Abbreviations are used for each modality: R $\rightarrow$ RGB, RD $\rightarrow$ RGBDiff, F $\rightarrow$ Flow, A $\rightarrow$ Audio. }
    \label{tab-sota}
    \vspace{-2mm}
    \centering
    \resizebox{\linewidth}{!}{
		\tablestyle{4pt}{1.15}
    \begin{tabular}{c|cccc|cc}
    \shline
    Method & Backbone & Input Size & Modality & Pre-train Data & UCF101 & HMDB51 \\ 
    \shline
    SpeedNet\#~\cite{benaim2020speednet} & S3D-G & $64 \times 224^2$ & R & K400 (28d) & 81.1 & 48.8 \\
    CoCLR*\#~\cite{han2020self} & S3D & $32 \times 128^2$ & R & K400 (28d) & 87.9 & 54.6 \\ 
    RSPNet*\#~\cite{chen2021rspnet} & S3D-G & $16 \times 224^2$ & R & K400 (28d) & 89.9 & 59.6 \\
    ASCNet*\#~\cite{huang2021ascnet} & S3D-G & $64 \times 224^2$ & R & K400 (28d) & 90.8 & 60.5 \\
    CoCLR*\#~\cite{han2020self} & S3D & $32 \times 128^2$ & R + F & K400 (28d) & 90.6 & 62.9 \\ 
    CVRL*\#~\cite{qian2021spatiotemporal} & R3D-50 & $32 \times 224^2$ & R & K400 (28d) & 92.2 & 66.7 \\ 
    \shline
    ST-Puzzle~\cite{kim2019self} & R3D-18 & $16 \times 80^2$ & R & K400 (28d) & 65.8 & 33.7 \\
    3D-RotNet~\cite{jing2018self} & R3D-18 & $16 \times 112^2$ & R & K600 (45d) & 66.0 & 37.1 \\
    DPC~\cite{han2019video} & R3D-18 & $25 \times 128^2$ & R & K400 (28d) & 68.2 & 34.5 \\
    RSPNet*~\cite{chen2021rspnet} & R3D-18 & $16 \times 112^2$ & R & K400 (28d) & 74.3 & 41.8 \\
    3D-RotNet~\cite{jing2018self} & R3D-18 & $16 \times 112^2$ & R + RD & K600 (45d) & 76.7 & 47.0 \\
    RTT~\cite{jenni2020video} & R3D-18 & $16 \times 112^2$ & R & K600 (45d) & 79.3 & 49.8 \\
    \hline
    TransRank-ST (\textbf{Ours}) & R3D-18 & $16 \times 112^2$ & R & K200 (9d) & 85.7 & 58.1 \\ 
    TransRank-ST (\textbf{Ours}) & R3D-18 & $16 \times 112^2$ & RD & K200 (9d) & 88.9 & 61.4 \\ 
    TransRank-ST (\textbf{Ours}) & R3D-18 & $16 \times 112^2$ & R + RD & UCF101 (1d) & 88.5 & 63.0 \\
    TransRank-ST (\textbf{Ours}) & R3D-18 & $16 \times 112^2$ & R + RD & SthV1 (4d) & 89.0 & 61.6 \\
    TransRank-ST (\textbf{Ours}) & R3D-18 & $16 \times 112^2$ & R + RD & K200 (9d) & \textbf{89.6} & \textbf{63.5} \\ 
    \shline
    VCOP~\cite{xu2019self} & R(2+1)D-18 & $16 \times 112^2$ & R & UCF101 (1d) & 72.4 & 30.9 \\
    PacePred*~\cite{wang2020self} & R(2+1)D-18 & $16 \times 112^2$ & R & K400 (28d)  & 77.1 & 36.6 \\ 
    VideoMoCo*~\cite{pan2021videomoco} & R(2+1)D-18 & $32 \times 112^2$ & R & K400 (28d) & 78.7 & 49.2 \\ 
    V3S~\cite{li2021video} & R(2+1)D-18 & $16 \times 112^2$ & R & K400 (28d) & 79.2 & 40.4 \\
    RSPNet*~\cite{chen2021rspnet} & R(2+1)D-18 & $16 \times 112^2$ & R & K400 (28d) & 81.1 & 44.6 \\
    RTT~\cite{jenni2020video} & R(2+1)D-18 & $16 \times 112^2$ & R & UCF101 (1d) & 81.6 & 46.4 \\
    XDC*~\cite{alwassel2019self} & R(2+1)D-18 & $32 \times 224^2$ & R + A & K400 (28d) & 86.8 & 52.6 \\
    \hline
    TransRank-ST (\textbf{Ours}) & R(2+1)D-18 & $16 \times 112^2$ & R & K200 (9d) & 87.8 & 60.1 \\ 
    TransRank-ST (\textbf{Ours}) & R(2+1)D-18 & $16 \times 112^2$ & D & K200 (9d) & 89.7 & 63.0 \\ 
    TransRank-ST (\textbf{Ours}) & R(2+1)D-18 & $16 \times 112^2$ & R + RD & K200 (9d) & \textbf{90.7} & \textbf{64.2} \\ 
    \shline  
    \end{tabular}}
    \vspace{-5mm}
\end{table}

\subsection{Comparison to the state-of-the-art}
We compare to prior works on self-supervised video representation learning in Table~\ref{tab-sota}.
We report Top-1 accuracies (average of 3 folds) on UCF101 and HMDB51 for both RGB and RGBDiff (the differences of RGB frames, cheap to obtain) modalities.
Since the backbones and pre-training settings used in prior works vary, we mainly compare results using commonly used network architectures (\emph{i.e.}, R3D-18, R(2+1)D-18). 
We adopt TransRank-ST with the T-Trans set $\{1\times, 2\times, \mathrm{rev}, \mathrm{rev}\mhyphen 2\times\}$ and two spatial tasks (the loss weight $\lambda$ is 0.5 for each task). 
We train TransRank-ST on all datasets for 200 epochs.

\textbf{Action Recognition. }
TransRank-ST achieves competitive performance with all backbones over both datasets. 
With a standard R3D-18 backbone, TransRank-ST outperforms the state-of-the-art RecogTrans approach RTT~\cite{jenni2020video} by 6.4\% and 8.3\% on UCF101 and HMDB51, with much fewer training data (K200 \vs K600). 
With RGB modality and the cheap RGBDiff modality combined, TransRank-ST with R(2+1)D-18 backbone achieves 90.7\% and 64.2\% Top-1 accuracies on UCF101 and HMDB51. 
TransRank outperforms the state-of-the-art RecogTrans methods by a large margin, and is comparable with state-of-the-art InstDisc approaches (\ie, CoCLR)  which adopts a more advanced backbone and a much longer input.

\textbf{Video Retrieval. }
We perform another quantitative evaluation of the learned representation via nearest-neighbor (NN) retrieval. 
We adopt TransRank-ST (R3D-18) trained on RGB inputs as the feature extractor. 
For each video, we extract and average features of 10 clips to obtain one single 512$\mhyphen$d feature vector. 
Cosine similarity is used as the metric to determine the nearest neighbors. 
For each clip in the testing split, we query clips in the training split and get $N$ nearest neighbors ($N=1, 5, 10$).
A query is correctly classified if $N$-nearest neighbors contain at least one video of the same class as the query.
We report the mean accuracy for different $N$ (R@N) and compare it to prior works in Table~\ref{tab-sota-retrieval}. 
TransRank-ST significantly outperforms all RecogTrans approaches, and achieves comparable performance to state-of-the-art InstDisc approaches like CoCLR~\cite{han2020self}.

\begin{table}[t]
  \vspace{-1mm}
  \captionsetup{font=small, position=top}
  \captionsetup[subfloat]{font=footnotesize, position=top}
  \caption{\textbf{Video retrieval performance on the split 1 of UCF101 and HMDB51. } 
    Three regions correspond to other RecogTrans approaches, TransRank-ST (\textbf{Ours}), 
    and InstDisc approaches. }
  \label{tab-sota-retrieval}
  \vspace{-2mm}
  \centering 
  \resizebox{\linewidth}{!}{
  \tablestyle{4pt}{1.2}
  \begin{tabular}{ccccccccc}
  \shline
  \multirow{2}{*}{Method} & \multirow{2}{*}{Backbone} & \multirow{2}{*}{\shortstack{Pre-train\\ Data}} & \multicolumn{3}{c}{UCF101} & \multicolumn{3}{c}{HMDB51} \\
  & & & R@1 & R@5 & R@10 & R@1 & R@5 & R@10 \\
  \shline 
  SpeedNet~\cite{benaim2020speednet} & S3D-G & K400 & 13.0 & 28.1 & 37.5 & - & - & - \\
  % VCOP~\cite{xu2019self} & R3D-18 & UCF101 & 14.1 & 30.3 & 40.0 & 7.6 & 22.9 & 33.4 \\
  VCP~\cite{luo2020video} & R3D-18 & UCF101 & 18.6 & 33.6 & 42.5 & 7.6 & 24.4 & 33.6 \\
  PRP~\cite{yao2020video} & R3D-18 & UCF101 & 22.8 & 38.5 & 46.7 & 8.2 & 25.8 & 38.5 \\
  RTT~\cite{jenni2020video} & R3D-18 & K600 & 26.1 & 48.5 & 59.1 & - & - & - \\
  V3S~\cite{li2021video} & R3D-18 & UCF101 & 28.3 & 43.7 & 51.3 & 10.8 & 30.6 & 42.3 \\
  \shline
  PacePred*~\cite{wang2020self} & R3D-18 & UCF101 & 23.8 & 38.1 & 46.4 & 9.6 & 26.9 & 41.1 \\
  RSPNet*~\cite{chen2021rspnet} & R3D-18 & K400 & 41.1 & 59.4 & 68.4 & - & - & - \\
  CoCLR*~\cite{han2020self} & S3D & UCF101 & 53.3 & 69.4 & 76.6 & 23.2 & 43.2 & 53.5 \\
  CoCLR*~\cite{han2020self} & S3D & K400 & 46.3 & 62.8 & 69.5 & 20.6 & 43.0 & 54.0 \\
  \shline
  TransRank-ST & R3D-18 & UCF101 & 46.5 & 63.7 & 72.8 & 19.4 & 45.4 & 59.1 \\
  TransRank-ST & R3D-18 & K200 & \textbf{54.0}& \textbf{71.8} & \textbf{79.6} & \textbf{25.5} & \textbf{52.3} & \textbf{65.8} \\
  \shline
  \end{tabular}}
\end{table}

\textbf{Comparisons to supervised pre-training. }
In Table~\ref{tab-cmp-supervised}, we compare TransRank-ST with supervised pre-training on 3 datasets: UCF101, HMDB51, SthV1, each with 3 settings: RGB, Diff, 2Stream. 
TransRank pre-train outperforms supervised pre-train across all downstream settings.

\begin{table}[t]
  \vspace{-2mm}
  \captionsetup{font=small, position=top}
  \captionsetup[subfloat]{font=footnotesize, position=top}
  \caption{\textbf{Comparisons with supervised pre-training. }}
  \label{tab-cmp-supervised}
  \vspace{-2mm}
  \centering 
  \resizebox{.95\linewidth}{!}{
  \tablestyle{8pt}{1.2}
  \begin{tabular}{ccccc}
  \shline
  Pre-training & Modality & UCF101 & HMDB51 & SthV1 \\
  \shline
  Supervised & RGB & 81.4 & 54.8 & 42.4 \\
  Supervised & Diff & 88.7 & 58.0 & 42.0 \\
  Supervised & 2Stream & 88.9 & 59.9 & 44.9 \\
  \shline
  TransRank-ST (\textbf{Ours}) & RGB & \textbf{85.7} & \textbf{58.1} & \textbf{43.1} \\
  TransRank-ST (\textbf{Ours}) & Diff & \textbf{88.9} & \textbf{61.4} & \textbf{43.2} \\
  TransRank-ST (\textbf{Ours}) & 2Stream & \textbf{89.6} & \textbf{63.5} & \textbf{45.8} \\
  \shline
  \end{tabular}}
  \vspace{-5mm}
\end{table}

\section{Discussion}
To conclude, we demonstrate the \textit{great potential} of \textit{RecogTrans}-based video self-supervised learning by introducing a unified framework named TransRank. 
We have shown its effectiveness through extensive experiments on ablation studies and comparisons with state-of-the-art methods. 
Given the initial success on marrying RecogTrans with InstDisc~\cite{wang2020self, chen2021rspnet, jenni2021time}, how to use TransRank to further boost this research line is also worth exploring.
We will release our code and pre-train models to facilitate future research.

\noindent
\textbf{Broader Impact.} 
Self-supervised learning is a data-hungry task, consuming expensive computational resources, though we have mitigated the effort and expense of collecting annotation. 
Since we have verified our model in multiple aspects and downstream tasks, we hope our released code and models can serve as a solid baseline for RecogTrans methods and deliver good initializations to benefit downstream tasks.
Besides, data-driven methods often bring the risk of learning biases and preserve them in downstream tasks. 
We encourage users to carefully consider the consequences of the biases when adopting our model.

\noindent
\textbf{Acknowledgement. }
This study is supported by the General Research Funds (GRF) of Hong Kong (No.14203518) and Shanghai Committee of Science and Technology, China (No. 20DZ1100800).

%% file: supp_downstream.tex
\begin{table}[t]
  \vspace{-2mm}
  \captionsetup{font=small, position=top}
  \captionsetup[subfloat]{font=footnotesize, position=top}
  \caption{\textbf{The mapping from HMDB51 action classes to five motion types. } 
  The mapping is manually annotated by ~\cite{parihar2021spatio}. }
  \label{tab-mhmdb}
  \vspace{-2mm}
  \centering 
  \resizebox{\linewidth}{!}{
  \tablestyle{4pt}{1.2}
  \begin{tabular}{c|c|c|c|c}
  \shline
  Linear & Projectile & Local & Oscillatory & Random \\
  \shline
  brush hair & cartwheel & chew & clap & draw sword \\
  climb & catch & drink & dribble & fall floor \\
  climb stairs & dive & eat & pullup & fencing \\
  punch & flic flac & kiss & pushup & hug \\
  push & golf & laugh & situp & kick \\
  ride bike & handstand & pour &  & kick ball \\
  ride horse & hit & shake hands &  & pick \\
  run & jump & shoot gun &  & sit \\
  shoot bow & shoot ball & smile &  & stand \\
  walk & somersault & smoke &  & sword \\
   & swing baseball & talk &  & sword exercise \\
   & throw & wave &  & turn \\
  \shline
  \end{tabular}}
\end{table}

\section{Details about Downstream Tasks}

\begin{figure}[t]
  \vspace{-2mm}
  \captionsetup{font=small, position=bottom}
  \captionsetup[subfloat]{font=footnotesize, position=bottom}
  \centering
  \subfloat[\textbf{Motion Type Prediction. }]{
    \includegraphics[width=.8\linewidth]{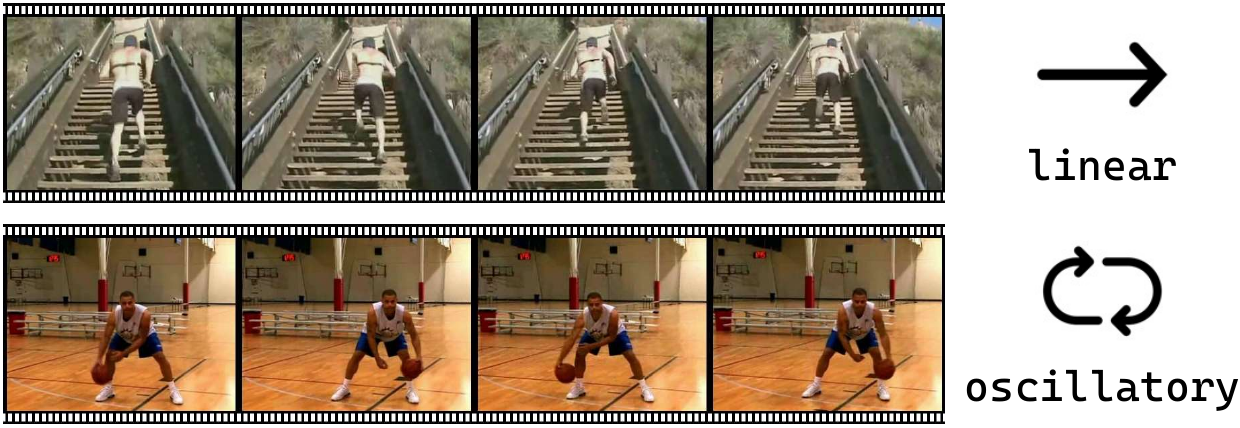}}
  \vspace{1mm}
  \centering
  \subfloat[\textbf{Clip Synchronization. }]{
    \includegraphics[width=.95\linewidth]{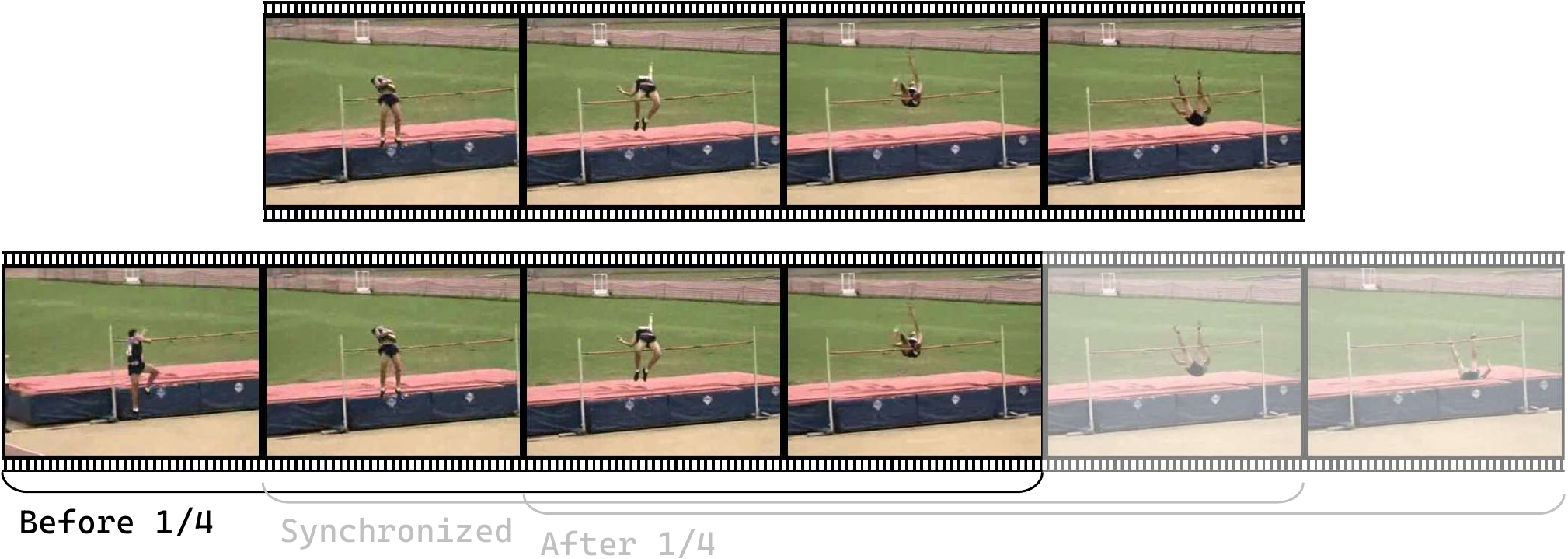}}
  \vspace{1mm}
  \centering
  \subfloat[\textbf{Temporal Order Prediction. }]{
    \includegraphics[width=.9\linewidth]{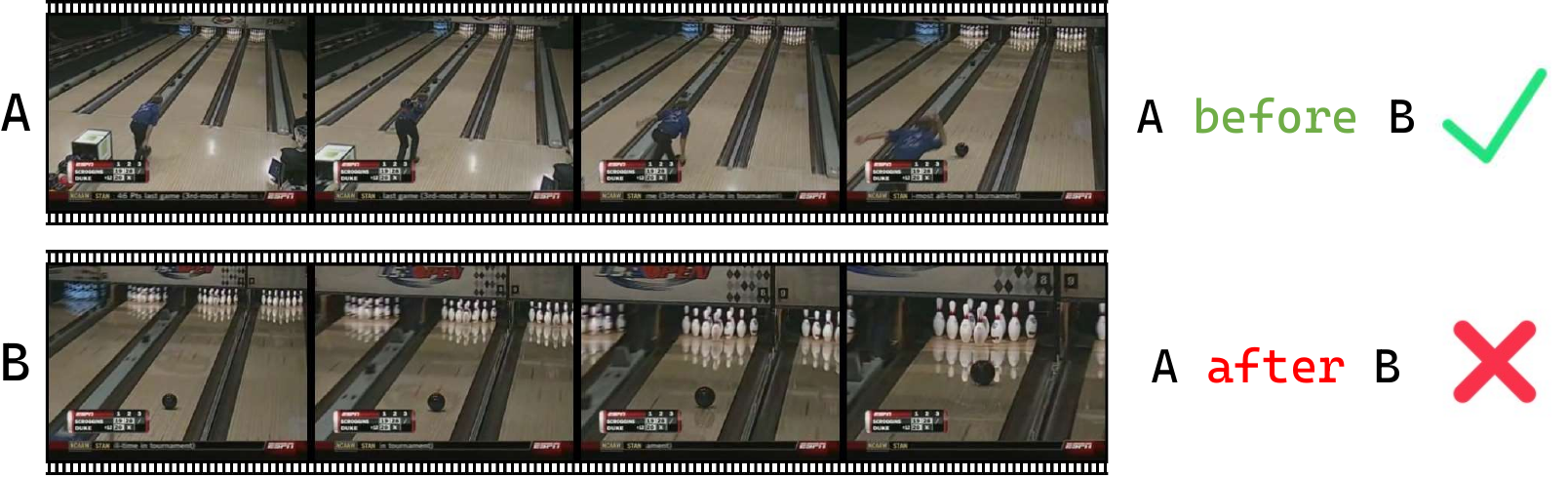}}
  \vspace{-2mm}
  \caption{\textbf{Three temporal-related downstream tasks. }}
  \label{fig-temporal-tasks}
  \vspace{-4mm}
\end{figure}

In this section, we first describe the three temporal-related downstream tasks in detail. 
Then we illustrate the concrete practices to transfer the SSL models to each task.
Figure~\ref{fig-temporal-tasks} illustrates the three temporal-related tasks.

\subsection{Temporal-related Tasks}

\textbf{Motion Type Prediction (\textit{Motion}). }
The task is first proposed in \cite{parihar2021spatio}, aiming at categorizing the motion in a video into five pre-defined categories: linear, projectile, local, oscillatory, and random.
To obtain the ground-truth, each class in HMDB51~\cite{kuehne2011hmdb} is annotated with one pre-defined motion type, yielding a new dataset mHMDB51~\cite{parihar2021spatio}.
We list the mapping in Table~\ref{tab-mhmdb}.
To transfer SSL models to this task, we initialize the backbone of a 5-way classifier with pre-trained weights and finetune on the mHMDB51. 
We report the Top1 accuracy on the test set (with 1530 videos). 

\begin{figure*}[t]
  \captionsetup{font=small, position=bottom}
  \centering
  \includegraphics[width=.85\linewidth]{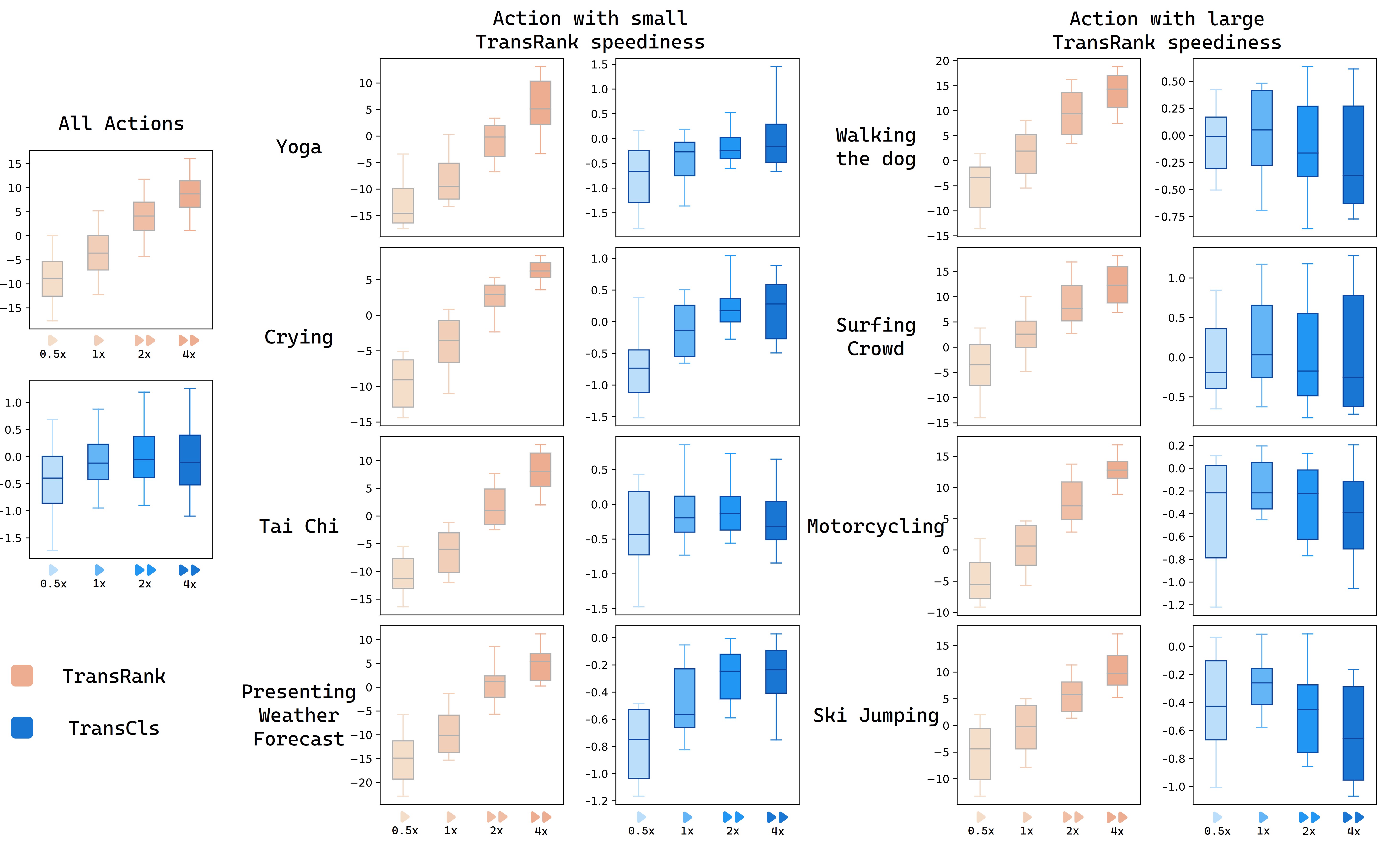}
  \vspace{-2mm}
  \caption{\textbf{Class-specific visualization for speediness score (unnormalized) distributions.} 
      TransRank can accurately capture the relative speediness of different playback rates, while TransCls cannot. 
      Besides, the absolute TransRank speediness score can reveal the speediness of an action category to some extent.
    }
  \label{fig-speediness-cs}
\end{figure*}

\textbf{Synchronization (\textit{Sync}). }
The synchronization task is first proposed in RTT~\cite{jenni2020video}, serving as a pseudo task to illustrate the motion modeling capability of SSL models. 
For this task, two temporally overlapping clips $a$, $b$ are sampled from a video and 
separately fed to the pre-trained backbone $\psi$ to extract features $\psi(a)$,  $\psi(b)$ at the $\mathrm{res_4}$ layer. 
We consider 7 overlapping patterns $\{-\nicefrac{3}{4}, -\nicefrac{1}{2}, ..., +\nicefrac{3}{4}\}$ 
(\eg, $-\nicefrac{1}{2}$ denotes $a$ is ahead of $b$ with $\nicefrac{1}{2}$ overlapping, $+\nicefrac{3}{4}$ denotes $a$ is behind of $b$ with $\nicefrac{3}{4}$ overlapping). 
Following~\cite{jenni2020video}, we adopt an MLP on top of the fused feature $\psi(a) - \psi(b)$ to recognize the overlapping pattern. 
The backbone parameters are fixed during training. 

\textbf{Temporal Order Prediction (\textit{Order}). } 
The order task is also proposed in RTT~\cite{jenni2020video} to evaluate the temporal modeling capability of SSL methods. 
For this task, a single input clip $x$ (with 16 frames) is constructed by sampling two non-overlapping sub-clips (each has 8 frames) $x_1, x_2$, where $x_1$ comes before $x_2$.
The network inputs are then either $(x_1, x_2)$ for class ``before'' or $(x_2, x_1)$ for class ``after''. 
Following~\cite{jenni2020video}, we train an MLP on top of the $\mathrm{res_4}$ feature of $x$.
The backbone parameters are fixed during training.

\subsection{Practices for Transfer Learning. }
In this section, we introduce the transfer learning practices we adopted for all pretext tasks in the preliminary study (main paper Sec~\red{3}). 
For all downstream experiments, we use SGD as the optimizer.
The hyper-parameter configuration is listed in Table~\ref{tab-hyper}.
We adopt five initial learning rates and report the best results. 

\textbf{Semantic-related Tasks. }
For \textit{Nearest Neighbor Evaluation, }
we extract and average features ($\mathrm{res_4}$) of 10 clips to obtain a single 512-d feature vector for each video. 
Cosine similarity is used as the metric to determine the nearest neighbors. 
For each clip in the testing split, we query clips in the training split to get N nearest neighbors (N = 1, 5, 10) and report the corresponding recalls.  
For \textit{Linear Evaluation} and \textit{Finetuning}, we adopt a fully-connected head on top of the backbone and re-train it for action recognition. 
We fix all backbone parameters for \textit{Linear Evaluation}.

\textbf{Temporal-related Tasks. }
We adopt the SSL weights to initialize the backbone and finetune all parameters for \textit{Motion}. 
For \textit{Sync} and \textit{Order}, we fix all backbone parameters and re-train the MLP heads. 
We adopt a 2-layer MLP with a hidden layer 512.

\begin{table}[t]
  \vspace{-2mm}
  \captionsetup{font=small, position=top}
  \captionsetup[subfloat]{font=footnotesize, position=top}
  \caption{\textbf{The hyper parameters for transfering. }}
  \label{tab-hyper}
  \vspace{-2mm}
  \centering 
  \resizebox{\linewidth}{!}{
  \tablestyle{12pt}{1.2}
  \begin{tabular}{cc}
  \shline
  Hyper Parameter & Value (Choices) \\
  \shline
  Batch Size & 128 \\
  Total Epochs & 100 \\
  Initial Learning Rate & $\{0.01, 0.02, 0.04, 0.08, 0.16\}$ \\
  Learning Rate Decay & 0.1, decays at the $60_{th}$, $80_{th}$ epoch \\
  Momentum & 0.9 \\ 
  Weight Decay & $10^{-4}$ \\
  Dropout Ratio & $0.5$ \\
  \shline 
  \end{tabular}}
  \vspace{-4mm}
\end{table}

%% file: supp_vis.tex
\section{Per-Class Speediness Distribution}

In Figure~\ref{fig-speediness-cs}, we visualize the distributions of speediness scores (unnormalized) for different action categories in MiniKinetics.
For comparison, we first visualize the unnormalized speediness distribution (TransRank and TransCls) for all MiniKinetics validation videos on the left side.
Figure~\ref{fig-speediness-cs} shows that for each single action category, 
For each action, TransRank can accurately capture the relative speediness of different playback rates, while TransCls cannot. 
Besides, the absolute TransRank speediness score can reveal some characteristics of the action.
The $1\times$ clips with lower TransRank speediness scores belong to the `still' actions, like Yoga, Crying, Tai Chi; 
while for actions like Motorcycling and Ski Jumping, the TransRank speediness scores of $1\times$ clips are much larger. 

%% file: supp_exp.tex
\begin{table*}[t]
  \vspace{-1mm}
  \captionsetup{font=small, position=top}
  \captionsetup[subfloat]{font=footnotesize, position=top}
  \caption{
    \textbf{Additional Results for Video Retrieval on the split 1 of UCF101 and HMDB51. } 
    For CoCLR, we report numbers obtained with the released checkpoint and codebase (\url{https://github.com/TengdaHan/CoCLR}). 
    For dual-modality video retrieval, we average the similarity of both modalities to obtain the new similarity matrix. 
    TransRank-ST achieves impressive retrieval performance with both RGB and the cheap RGBDiff modalities. 
    With two modalities combined, TransRank-ST outperforms the previous state-of-the-art CoCLR, which adopts the much more expensive modality optical flow. }
  \label{tab-sota-retrieval}
  \vspace{-2mm}
  \centering 
  \resizebox{\linewidth}{!}{
  \tablestyle{7pt}{1.4}
  \begin{tabular}{cccccccccccccc}
  \shline
  \multirow{2}{*}{Method} & \multirow{2}{*}{Backbone} & \multirow{2}{*}{Modality} & \multirow{2}{*}{\shortstack{Pre-train\\ Data}} & \multicolumn{5}{c}{UCF101} & \multicolumn{5}{c}{HMDB51} \\
  & & & & R@1 & R@5 & R@10 & R@20 & R@50 & R@1 & R@5 & R@10 & R@20 & R@50 \\
  \shline
  CoCLR & S3D & RGB & UCF101 & 53.2 & 69.2 & 76.5 & 82.2 & 88.8 & 21.9 & 42.1 & 54.4 & 68.0 & 83.9 \\
  CoCLR & S3D & Flow & UCF101 & 49.2 & 68.2 & 75.7 & 82.0 & 88.4 & 22.2 & 46.1 & 56.1 & 69.3 & 84.1 \\
  CoCLR & S3D & RGB + Flow & UCF101 & 54.5 & 71.2 & 76.8 & 82.6 & 89.0 & 23.6 & 46.6 & 57.9 & 70.1 & 85.0 \\
  \hline 
  CoCLR & S3D & RGB & K400 & 46.3 & 62.8 & 69.5 & 76.7 & 84.5 & 20.6 & 43.0 & 54.0 & 66.3 & 81.2 \\
  CoCLR & S3D & Flow & K400 & 23.7 & 46.1 & 58.1 & 70.1 & 82.8 & 11.4 & 33.1 & 48.0 & 64.3 & 84.3 \\
  CoCLR & S3D & RGB + Flow & K400 & 40.3 & 60.6 & 69.5 & 77.6 & 86.9 & 19.3 & 43.6 & 54.3 & 68.9 & 84.8 \\
  \shline 
  TransRank-ST & R3D-18 & RGB & UCF101 & 46.5 & 63.7 & 72.8 & 82.0 & 90.0 & 19.4 & 45.4 & 59.1 & 74.0 & 86.9 \\
  TransRank-ST & R3D-18 & RGBDiff & UCF101 & 43.7 & 62.7 & 72.8 & 82.2 & 91.2 & 17.6 & 41.2 & 56.4 & 71.3 & 87.1 \\
  TransRank-ST & R3D-18 & RGB + Diff & UCF101 & 48.1 & 66.2 & 75.0 & 83.0 & 91.5 & 19.7 & 47.2 & 60.1 & 74.0 & 86.6 \\
  \hline
  TransRank-ST & R3D-18 & RGB & K200 & 54.0 & 71.8 & 79.6 & 86.4 & 92.5 & 25.5 & \textbf{52.3} & 65.8 & \textbf{78.4} & 89.6  \\
  TransRank-ST & R3D-18 & RGBDiff & K200 & 52.9 & 72.7 & 81.6 & 87.6 & 93.6 & 21.8 & 50.0 & 62.8 & 75.4 & \textbf{90.9} \\
  TransRank-ST & R3D-18 & RGB + Diff & K200 & \textbf{56.7} & \textbf{74.2} & \textbf{82.1} & \textbf{88.3} & \textbf{93.9} & \textbf{27.3} & 52.1 & \textbf{66.9} & 77.7 & 90.6 \\
  \shline
  \end{tabular}}
\end{table*}

\section{Experiments }

\subsection{Training Details.}
\textbf{Pre-training Details. }
In pre-training, we sample $N$ clips from each video, applying different temporal transformations ($1\times$, $2\times$, $\mathrm{rev}$, \etc), spatial transformations (like RandomRotate, only for TransRank-ST), and strong clip-wise spatial \& temporal augmentations to each clip. 
With transformations applied, each input clip consists of 16 frames with spatial size $112\times 112$. 
We use SGD as the optimizer with a mini-batch size 64. 
We adopt the CosineAnnealing scheduler to update the learning rate (lr), while the initial lr is set to 0.1. 
We train the model for 100 epochs by default. 

\textbf{Finetuning Details. }
We finetune TransRank on UCF101, HMDB51, and SthV1 for action recognition. 
We finetune 100 epochs on UCF101, HMDB51; 50 epochs on SthV1 (the dataset is much larger). 
SGD is used as the optimizer with the MultiStepLR scheduler (decay the learning rate by $\nicefrac{1}{10}$ after $\nicefrac{3}{5}$ and $\nicefrac{4}{5}$ training epochs finished).
By default, we use 0.16 as the finetuning LR and set the batch size to 128. 
We find that large finetuning LR is critical for the success of RecogTrans-based SSL approaches.

\subsection{Multi-modality Retrieval Results}

In Table~\ref{tab-sota-retrieval}, we report multi-modality video retrieval results. 
Besides RGB, TransRank-ST can also achieve impressive retrieval performance with the RGBDiff modality. 
Moreover, we find that RGB and RGBDiff are two complementary modalities. 
The ensemble can outperform each individual modality. 
Integrated with the cheap modality RGBDiff, 
TransRank-ST outperforms the contrastive-based approach CoCLR~\cite{han2020self} trained with the much more expensive modality optical flow. 
% TransRank-ST pretrained on K200 surpasses CoCLR trained on either UCF101 or K400, while 
Besides, we find a clear trend for TransRank-ST: more pre-training data $\rightarrow$ better retrieval performance. 
However, this trend doesn't hold true for the contrastive-based CoCLR. 

\subsection{Qualitative Results for Video Retrieval}

\begin{figure*}[t]
  \vspace{-4mm}
  \captionsetup{font=small, position=bottom}
  \centering
  \includegraphics[width=\linewidth]{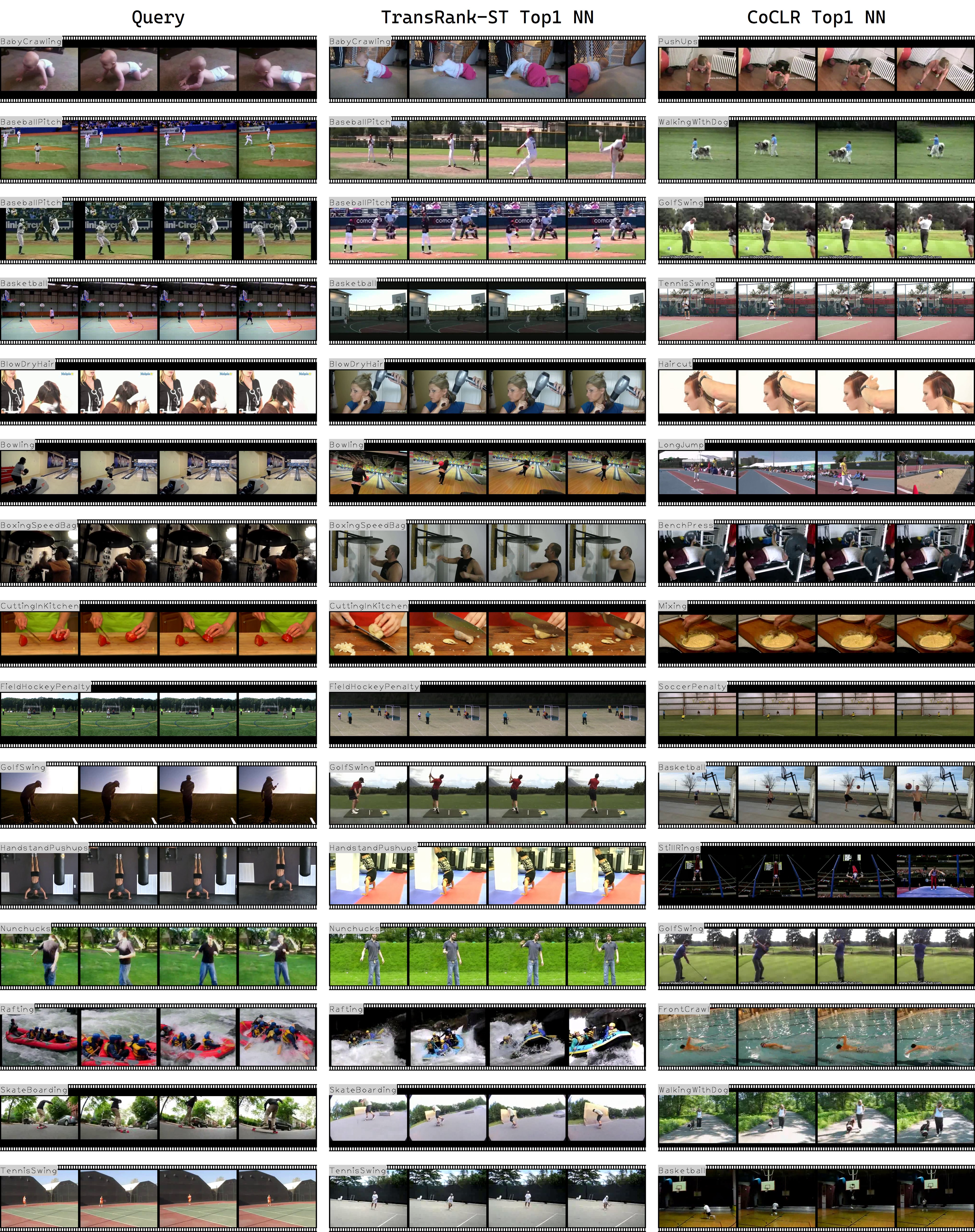}
  \caption{\textbf{Qualitative results for video retrieval.} 
    The representation learned by TransRank-ST can retrieve videos with the same action categories.
    It focuses on human motion and is less vulnerable to changes in background scene or illumination. 
  }
  \label{fig-retrieval-vis}
\end{figure*}

Figure~\ref{fig-retrieval-vis} visualizes a query video clip and its Top1 retrievals obtained with both TransRank-ST and CoCLR (both use RGB modality).
We find that compared to CoCLR, TransRank-ST focuses less on static cues and more on human motions.
With TransRank-ST features, one can obtain high-quality retrieval results, robust to changes in background scene or illumination.

